\documentclass{article} 
\usepackage[final]{nips_2016_arxiv}
\usepackage{hyperref}
\usepackage{url}
\usepackage{textcomp}
\usepackage{floatrow}
\usepackage{graphicx}
\usepackage{amsmath}
\usepackage{amsfonts}
\usepackage{todonotes}
\usepackage{subfigure}
\usepackage{epstopdf}
\usepackage{tabulary}
\usepackage{multirow}
\usepackage{color}
\usepackage{booktabs}
\newfloatcommand{capbtabbox}{table}[][\FBwidth]

\graphicspath{ {figures/}}

\title{Using Fast Weights to Attend to the Recent Past}

\author{
Jimmy Ba \\
University of Toronto \\
{\small \texttt{jimmy@psi.toronto.edu}} \\
\And
Geoffrey Hinton \\
University of Toronto and Google Brain\\
{\small \texttt{geoffhinton@google.com}} \\
\AND
Volodymyr Mnih \\
Google DeepMind \\
{\small \texttt{vmnih@google.com} } \\
\And
Joel Z. Leibo \\
Google DeepMind \\
{\small \texttt{jzl@google.com} } \\
\And
Catalin Ionescu \\
Google DeepMind \\
{\small \texttt{cdi@google.com} } \\
}

\newcommand{\cut}[1]{}

\newcommand{\be}{\begin{equation}}
\newcommand{\ee}{\end{equation}}
\def\bea#1\eea{\begin{align}#1\end{align}}
\def\bean#1\eean{\begin{align*}#1\end{align*}}
\makeatletter
\renewcommand*\env@matrix[1][\arraystretch]{%
  \edef\arraystretch{#1}%
  \hskip -\arraycolsep
  \let\@ifnextchar\new@ifnextchar
  \array{*\c@MaxMatrixCols c}}
\makeatother

\begin{document}

\maketitle
\vspace{-0.2in}
\begin{abstract}
Until recently, research on artificial neural networks was largely restricted to systems with only two types of variable: Neural activities that represent the current or recent input and weights that learn to capture regularities among inputs, outputs and payoffs. There is no good reason for this restriction. Synapses have dynamics at many different time-scales and this suggests that artificial neural networks might benefit from variables that change slower than activities but much faster than the standard weights.  These ``fast weights'' can be used to store temporary memories of the recent past and they provide a neurally plausible way of implementing the type of attention to the past that has recently proved very helpful in sequence-to-sequence models. By using fast weights we can avoid the need to store copies of neural activity patterns.  
\end{abstract}

\section{Introduction}
\label{sec:intro}

Ordinary recurrent neural networks typically have two types of memory that have very different time scales, very different capacities and very different computational roles.  The history of the sequence currently being processed is stored in the hidden activity vector, which acts as a short-term memory that is updated at every time step.  The capacity of this memory is ${\rm O} (H)$ where $H$ is the number of hidden units. Long-term memory about how to convert the current input and hidden vectors into the next hidden vector and a predicted output vector  is stored in the weight matrices connecting the hidden units to themselves and to the inputs and outputs. These matrices are typically updated at the end of a sequence and their capacity is ${\rm O}(H^2) + {\rm O} (IH) + {\rm O}(HO)$ where $I$ and $O$ are the numbers of input and output units.  

Long short-term memory networks \citep{hochreiter1997long} are a more complicated type of RNN that work better for discovering long-range structure in sequences for two main reasons: First, they compute {\it increments} to the hidden activity vector at each time step rather than recomputing the full vector\footnote{This assumes the ``remember gates '' of the LSTM memory cells are set to one.}. This encourages  information in the hidden states to persist for much longer. Second, they allow the hidden activities to determine the states of gates that scale the effects of the weights. These multiplicative interactions allow the effective weights to be dynamically adjusted by the input or hidden activities via the gates. However, LSTMs are still limited to a short-term memory capacity of ${\rm O}(H)$ for the history of the current sequence. 

Until recently, there was surprisingly little practical investigation of other forms of memory in recurrent nets despite strong psychological evidence that it exists and obvious computational reasons why it was needed. There were occasional suggestions that neural networks could benefit from a third form of memory that has much higher storage capacity than the neural activities but much faster dynamics than the standard “slow” weights. This memory could store information specific to the history of the current sequence so that this information is available to influence the ongoing processing without using up the memory capacity of the hidden activities. \citet{hinton1987using} suggested that fast weights could be used to allow true recursion in a neural network and \citet{schmidhuber1993reducing} pointed out that a system of this kind could be trained end-to-end using backpropagation, but neither of these papers actually implemented this method of achieving recursion. 

\section{Evidence from physiology that temporary memory may not be stored as neural activities} 

Processes like working memory, attention, and priming operate on timescale of 100ms to minutes. This is simultaneously too slow to be mediated by neural activations without dynamical attractor states (10ms timescale) and too fast for long-term synaptic plasticity mechanisms to kick in (minutes to hours). While artificial neural network research has typically focused on methods to maintain temporary state in activation dynamics, that focus may be inconsistent with evidence that the brain also---or perhaps primarily---maintains temporary state information by short-term synaptic plasticity mechanisms \citep{tsodyks1998neural,  abbott2004synaptic, barak2007persistent}.

The brain implements a variety of short-term plasticity mechanisms that operate on intermediate timescale. For example, short term facilitation is implemented by leftover  $[\text{Ca}^{2+}]$ in the axon terminal after depolarization while short term depression is implemented by presynaptic neurotransmitter depletion \cite{zucker2002short}. Spike-time dependent plasticity can also be invoked on this timescale \citep{markram1997regulation, bi1998synaptic}. These plasticity mechanisms are all synapse-specific. Thus they are more accurately modeled by a memory with $O(H^2)$ capacity than the $O(H)$ of standard recurrent artificial recurrent neural nets and LSTMs. 

\section{Fast Associative Memory}

One of the main preoccupations of neural network research in the 1970s and early 1980s \citep{willshaw1969non, kohonen1972correlation, anderson1981models, hopfield1982neural} was the idea that memories were not stored by somehow keeping copies of patterns of neural activity.  Instead, these patterns  were reconstructed when needed from information stored in the weights of an associative network and the very same weights could store many different memories
An auto-associative memory that has $N^2$ weights cannot be expected to store more that N real-valued vectors with N components each.   How close we can come to this upper bound depends on which storage rule we use. 
Hopfield nets use a simple, one-shot, outer-product storage rule and achieve a capacity of approximately $0.15N$ binary vectors using weights that require $\text{log}(N)$ bits each.  Much more efficient use can be made of the weights by using an iterative, error correction storage rule to learn weights that can retrieve each bit of a pattern from all the other bits \citep{gardner1988space}, but for our purposes maximizing the capacity is less important than having a simple, non-iterative storage rule, so we will use an outer product rule to store hidden activity vectors in fast weights that decay rapidly.  The usual weights in an RNN will be called slow weights and they will learn by stochastic gradient descent in an objective function taking into account the fact that changes in the slow weights will lead to changes in what gets stored automatically in the fast associative memory.

A fast associative memory has several advantages when compared with the type of memory assumed by a Neural Turing Machine (NTM) \citep{graves2014neural}, Neural Stack \citep{grefenstette2015learning}, or Memory Network \citep{weston2014memory}. First, it is not at all clear how a real brain would implement the more exotic structures in these models e.g., the tape of the NTM, whereas it is clear that the brain could implement a fast associative memory in synapses with the appropriate dynamics. Second, in a fast associative memory there is no need to decide where or when to write to memory and where or when to read from memory. The fast memory is updated all the time and the writes are all superimposed on the same fast changing component of the strength of each synapse. Every time the input changes there is a transition to a new hidden state which is determined by a combination of three sources of information: The new input via the slow input-to-hidden weights, $C$, the previous hidden state via the slow transition weights, $W$, and the recent history of hidden state vectors via the fast weights, $A$.  The effect of the first two sources of information on the new hidden state can be computed once and then maintained as a sustained boundary condition for a brief iterative settling process which allows the fast weights to influence the new hidden state. Assuming that the fast weights decay exponentially, we now show that the effect of the fast weights on the hidden vector during an iterative settling phase is to provide an additional input that is proportional to the sum over all recent hidden activity vectors of the scalar product of that recent hidden vector with the current hidden activity vector, with each term in this sum being weighted by the decay rate raised to the power of how long ago that hidden vector occurred. So fast weights act like a kind of attention to the recent past but with the strength of the attention being determined by the scalar product between the current hidden vector and the earlier hidden vector rather than being determined by a separate parameterized computation of the type used in neural machine translation models \citep{neural-machine-translation}.  

\begin{figure}[t]
    \centering
    \vspace{-0.7in}
    \includegraphics[height=3.2in]{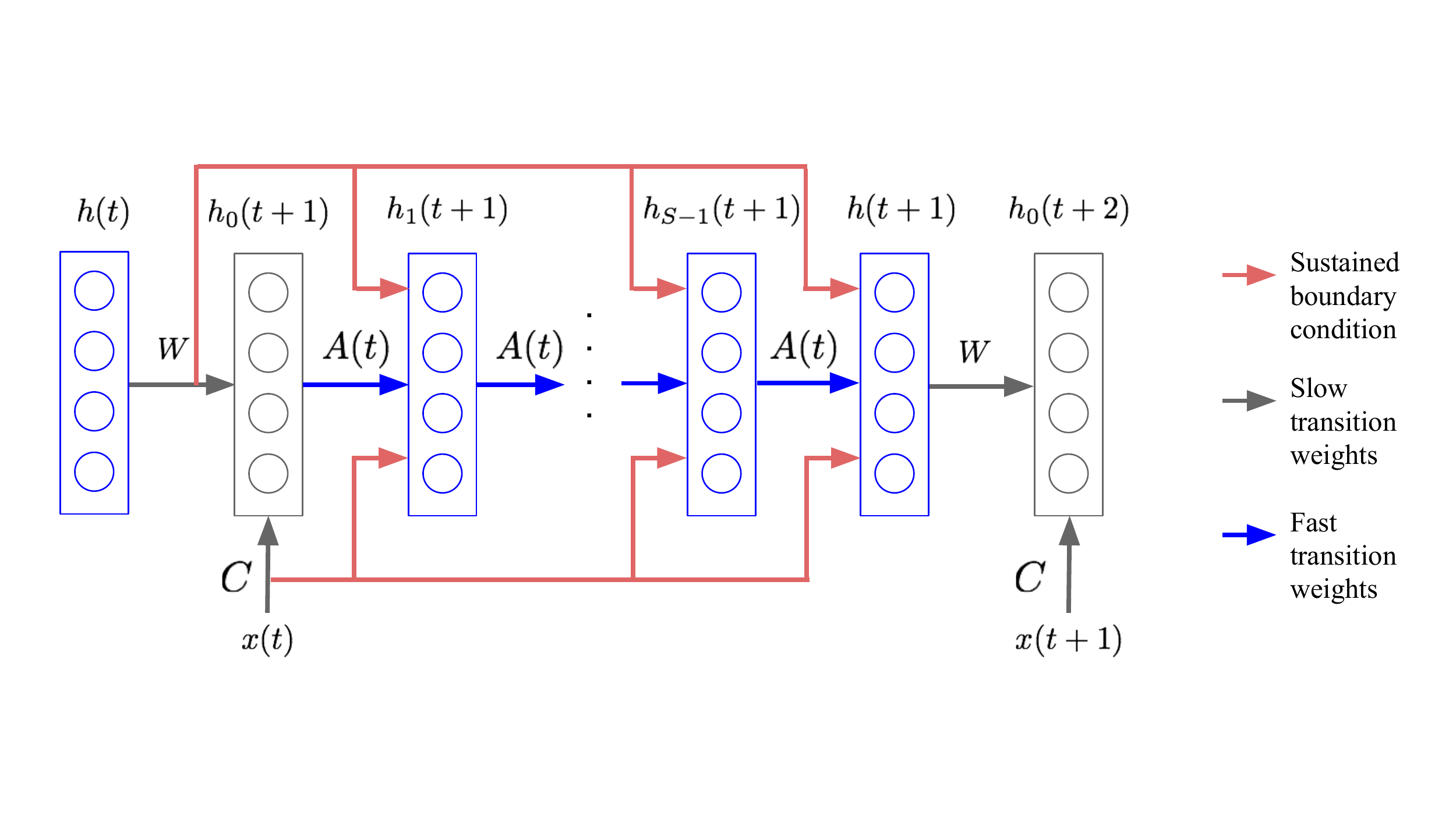}
    ~\\[-0.8in]
    \caption{\label{fig:model} The fast associative memory model.}
    
\end{figure}

The update rule for the fast memory weight matrix, $A$, is simply to multiply the current fast weights by a decay rate, $\lambda$, and add the outer product of the hidden state vector, $h(t)$, multiplied by a learning rate, $\eta$: 
\begin{equation}
A(t) = \lambda A(t-1) + \eta h(t) h(t)^T 
\end{equation}

The next vector of hidden activities, $h(t+1)$, is computed in two steps.  The ``preliminary'' vector $h_0(t+1)$ is determined by the combined effects of the input vector $x(t)$ and the previous hidden vector:  $h_0(t+1) = f(W h(t) + C x(t)) $, where $W$ and $C$ are slow weight matrices and $f(.)$ is the nonlinearity used by the hidden units. The preliminary vector is then used to initiate an ``inner loop''  iterative process which runs for S steps and progressively changes the hidden state into $h(t+1) = h_S(t+1)$
\begin{equation}
  h_{s+1}(t+1) =  f([W h(t) + C x(t)] + A(t) h_s(t+1)),
\end{equation}

where the terms in square brackets are the sustained boundary conditions. In a real neural net, $A$ could be implemented by rapidly changing synapses
but in a computer simulation that uses sequences which have fewer time steps than the dimensionality of $h$, $A$ will be of less than full rank and it is more efficient to compute the term $A(t) h_s(t+1)$ without ever computing the full fast weight matrix, $A$. Assuming $A$ is $0$ at the beginning of the sequence, 
\begin{equation}
A(t) = \eta \sum_{\tau = 1}^{\tau = t} \lambda^{t-\tau}h(\tau)h(\tau)^T
\end{equation}
\begin{equation}
  A(t) h_{s}(t+1)  =  \eta \sum_{\tau = 1}^{\tau = t} \lambda^{t-\tau} h(\tau) [h(\tau)^T h_{s}(t+1)]\label{eq:fw_lr}
\end{equation}

The term in square brackets is just the scalar product of an earlier hidden state vector, $h(\tau)$, with the current hidden state vector,  $h_{s}(t+1)$, during the iterative inner loop. So at each iteration of the inner loop, the fast weight matrix is exactly equivalent to attending to past hidden vectors in proportion to their scalar product with the current hidden vector, weighted by a decay factor. During the inner loop iterations, attention will become more focussed on past hidden states that manage to attract the current hidden state. 

The equivalence between using a fast weight matrix and comparing with a set of stored hidden state vectors is very helpful for computer simulations.  It allows us to explore what can be done with fast weights without incurring the huge penalty of having to abandon the use of mini-batches during training. At first sight, mini-batches cannot be used because the fast weight matrix is different for every sequence, but comparing with a set of stored hidden vectors does allow mini-batches. 

\subsection{Layer normalized fast weights}
A potential problem with fast associative memory is that the scalar product of two hidden vectors could vanish or explode depending on the norm of the hidden vectors.  Recently, layer normalization \citep{ln} has been shown to be very effective at stablizing the hidden state dynamics in RNNs and reducing training time.  Layer normalization is applied to the vector of summed inputs to all the recurrent units at a particular time step. It uses the mean and variance of the components of this vector to re-center and re-scale  those summed inputs. Then,  before applying the nonlinearity,  it includes a learned, neuron-specific bias and gain.  We apply layer normalization to the fast associative memory as follows: 
\bea h_{s+1}(t+1) = f(\mathcal{LN}[W h(t) + C x(t) +A(t) h_s(t+1)]) \eea 
where $\mathcal{LN}[.]$ denotes layer normalization. We found that applying layer normalization on each iteration of the inner loop makes the fast associative memory more robust to the choice of learning rate and decay hyper-parameters. For the rest of the paper, fast weight models are trained using layer normalization and the outer product learning rule with fast learning rate of 0.5 and decay rate of 0.95, unless otherwise noted.

\section{Experimental results}
To demonstrate the effectiveness of the fast associative memory, we first investigated the problems of associative retrieval (section \ref{section:retrieval}) and MNIST classification (section \ref{section:mnist}). We compared fast weight models to regular RNNs and LSTM variants. We then applied the proposed fast weights to a facial expression recognition task using a  fast associative memory model to store the results of processing at one level while examining a sequence of details at a finer level (section \ref{section:faces}). The hyper-parameters of the experiments were selected through grid search on the validation set. All the models were trained using mini-batches of size 128 and the Adam optimizer \citep{adam}. A description of the training protocols and the hyper-parameter settings we used can be found in the Appendix. Lastly, we show that fast weights can also be used effectively to implement reinforcement learning agents with memory (section \ref{section:agents}). 

\subsection{Associative retrieval} \label{section:retrieval}
We start by demonstrating that the method we propose for storing and retrieving temporary memories works effectively for a toy task to which it is very well suited.  Consider a task where multiple key-value pairs are presented in a sequence. At the end of the sequence, one of the keys is presented and the model must predict the value that was temporarily associated with the key.  We used strings that contained characters from English alphabet, together with the digits 0 to 9. To construct a training sequence, we first randomly sample a character from the alphabet without replacement. This is the first key. Then a single digit is sampled as the associated value for that key.  After generating a sequence of $K$ character-digit pairs, one of the $K$ different characters is selected at random as the query and the network must predict the associated digit. Some examples of such string sequences and their targets are shown below:
\begin{align}
  \text{Input string}& \text{\ \  Target} \nonumber  \\
  \text{c9k8j3f1??c}& \text{\ \ \ \ 9} \nonumber  \\
  \text{j0a5s5z2??a}& \text{\ \ \ \ 5} \nonumber
\end{align}
where `?' is the token to separate the query from the key-value pairs. We generated 100,000 training examples, 10,000 validation examples and 20,000 test examples.  To solve this task, a standard RNN has to end up with hidden activities that somehow store all of the key-value pairs after the keys and values are presented sequentially. This makes it a significant challenge for models only using slow weights.

We used a neural network with a single recurrent layer for this experiment. The recurrent network processes the input sequence one character at a time. The input character is first converted into a learned 100-dimensional embedding vector which then provides input to the recurrent layer\footnote{To make the architecture for this task more similar to the architecture for the next task we first compute a 50 dimensional embedding vector and then expand this to a 100-dimensional embedding.}.  The output of the recurrent layer at the end of the sequence is then processed by another hidden layer of 100 ReLUs before the final softmax layer.  We augment the ReLU RNN with a fast associative memory and compare it to an LSTM model with the same architecture. Although the original LSTMs do not have explicit long-term storage capacity, recent work from \cite{danihelka2016associative} extended LSTMs by adding complex associative memory. In our experiments, we compared fast associative memory to both LSTM variants.

\begin{figure}
\vspace{-0.2in}
\begin{floatrow}
\capbtabbox{%
        \begin{tabular}{llll}
       	\toprule
        Model & R=20& R=50 & R=100 \\ 
        \toprule
		IRNN & 62.11\%   & 60.23\% & 0.34\% \\
		\midrule
		LSTM & 60.81\%   & 1.85\%  & 0\% \\
		\midrule
		A-LSTM& 60.13\%  & 1.62\%  & 0\% \\
		\midrule
		Fast weights & 1.81\% & 0\%& 0\% \\
		\bottomrule
\end{tabular}
}{%
 \caption{ \label{table:string} \small Classification error rate comparison on the associative retrieval task. }
} 
\ffigbox{%
    \centering
    \vspace{-0.in}
    \includegraphics[width=5.8cm]{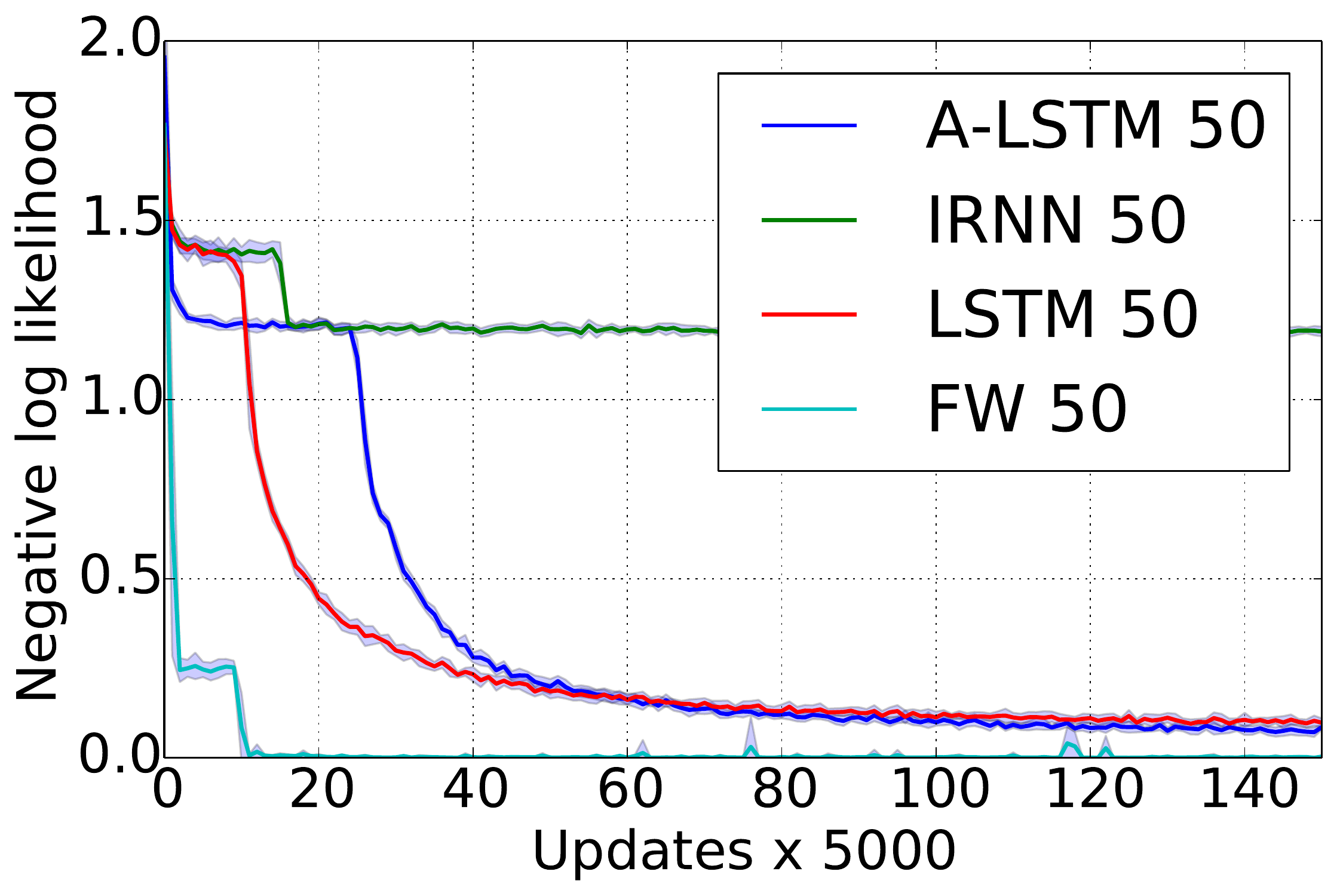}
    \vspace{-0.1in}
}
{%
  \caption{ \label{fig:string} \small Comparison of the test log likelihood on the associative retrieval task with 50 recurrent hidden units.} 
    }
\end{floatrow}
\end{figure}

Figure~\ref{fig:string} and Table~\ref{table:string} show that when the number of recurrent units is small, the fast associative memory significantly outperforms the LSTMs with the same number of recurrent units. The result fits with our hypothesis that the fast associative memory allows the RNN to use its recurrent units more effectively.  In addition to having higher retrieval accuracy, the model with fast weights also converges faster than the  LSTM models. 

\subsection{Integrating glimpses in visual attention models}\label{section:mnist}

Despite their many successes, convolutional neural networks are computationally expensive and the representations they learn can be hard to interpret. Recently, visual attention models \citep{recurrent-attention, dram, xu2015show} have been shown to overcome some of the limitations in ConvNets. One can understand what signals the algorithm is using by seeing where the model is looking. Also, the visual attention model is able to selectively focus on important parts of visual space and thus avoid any detailed processing of much of the background clutter.  In this section, we show that visual attention models can use fast weights to store information about object parts, though we use a very restricted set of glimpses that do not correspond to natural parts of the objects. 

Given an input image, a visual attention model computes a sequence of glimpses over regions of the image. The model not only has to determine where to look next, but also has to remember what it has seen so far in its working memory so that it can make  the correct classification later. Visual attention models can learn to find multiple objects in a large static input image and classify them correctly, but the learnt glimpse policies are typically over-simplistic: They only use a single scale of glimpses and they tend to scan over the image in a rigid way.  Human eye movements and fixations are far more complex. The ability to focus on different parts of a whole object at different scales allows humans to apply the very same knowledge in the weights of the network at many different scales, but it requires some form of temporary memory  to allow the network to integrate what it discovered in a set of glimpses.  Improving the model's ability to remember recent glimpses should help the visual attention model to discover non-trivial glimpse policies. Because the fast weights can store all the glimpse information in the sequence, the hidden activity vector is freed up to learn how to intelligently integrate visual information and retrieve the appropriate memory content for the final classifier. 

To explicitly verify that larger memory capacity is beneficial to visual attention-based models, we simplify the learning process in the following way: First, we provide a pre-defined glimpse control signal so the model knows where to attend rather than having to learn the control policy through reinforcement learning.  Second, we introduce an additional control signal to the memory cells so the attention model knows when to store the glimpse information. A typical visual attention model is complex and has high variance in its performance due to the need to learn the policy network and the classifier at the same time. Our simplified learning procedure enables us to discern the performance improvement contributed by using fast weights to remember the recent past.

We consider a simple recurrent visual attention model that has 
a similar architecture to the RNN from the previous experiment. It does not predict where to attend but rather is given a fixed sequence of locations: the static input image is broken down into four non-overlapping quadrants recursively with two scale levels. The four coarse regions, down-sampled to $7\times 7$,  along with their the four $7\times 7$ quadrants are presented in a single sequence as shown in Figure~\ref{fig:model}. Notice that the two glimpse scales form a two-level hierarchy in the visual space.  In order to solve this task successfully, the attention model needs to integrate the glimpse information from different levels of the hierarchy. One solution is to use the model's hidden states to both store and integrate the glimpses of different scales. A much more efficient solution is to use a temporary ``cache'' to store any of the unfinished glimpse computation when processing the glimpses from a finer scale in the hierarchy. Once the computation is finished at that scale, the results can be integrated with the partial results at the higher level by ``popping'' the previous result from the ``cache''. Fast weights, therefore, can act as a neurally plausible ``cache'' for storing partial results. The slow weights of the same model can then specialize in integrating glimpses at the same scale. Because the slow weights are shared for all glimpse scales, the model should be able to store the partial results at several levels in the same set of fast weights, though we have only demonstrated the use of fast weights for storage at a single level. 

\begin{figure}[t]
    \label{fig:rfw}
    \centering
    \vspace{-0.5in}
    \includegraphics[height=3.2in]{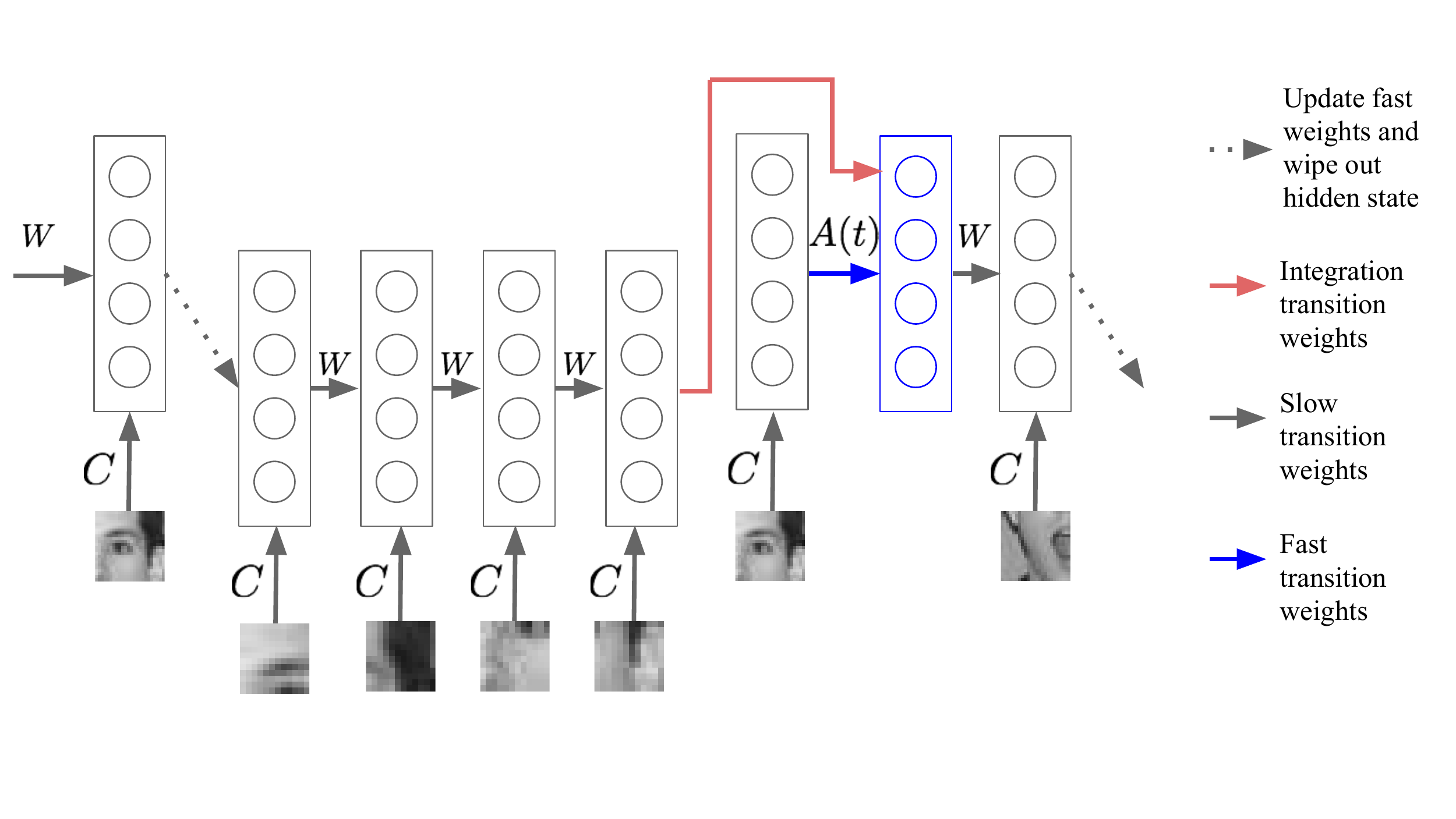}
    ~\\[-.8in]
    \caption{\label{fig:rmodel} The multi-level fast associative memory model.}
    \vspace{0.2in}
\end{figure}
\begin{table}
\begin{tabular}{llll}
    \toprule
        Model & 50 features & 100 features & 200 features \\ 
    \toprule
        IRNN  & 12.95\% & 1.95\% & 1.42\% \\
    \midrule
        LSTM & 12\% & 1.55\% & 1.10\% \\
    \midrule
        ConvNet & \bf 1.81\% & \bf 1.00\% & \bf 0.9\% \\
    \midrule
    Fast weights & 7.21\% & 1.30\% & \bf 0.85\%\\
    \bottomrule
\end{tabular}
\caption{ \label{table:mnist} \small Classification error rates on MNIST.}
\end{table}
We evaluated the multi-level visual attention model on the MNIST handwritten
digit dataset. MNIST is a well-studied
problem on which many other techniques have been benchmarked. It
contains the ten classes of handwritten digits, ranging from 0 to 9.
The task is to predict the class label of an isolated and roughly normalized  28x28 image of a digit. The glimpse sequence, in this case, consists of 24 patches of
$7\times 7$ pixels. 

Table~\ref{table:mnist} compares classification results for a ReLU RNN with a multi-level fast associative memory against an LSTM that gets the same sequence of glimpses.  Again the result shows that when the number of hidden units is limited,  fast weights give a significant improvement over the other models. As we increase the memory capacities, the multi-level fast associative memory consistently outperforms the LSTM in classification accuracy. 

Unlike models that must integrate a sequence of glimpses,  convolutional neural networks process all the glimpses in parallel and use layers of hidden units to hold all their intermediate computational results.  We further demonstrate the effectiveness of the fast weights by comparing to a three-layer convolutional neural network that uses the same patches as the glimpses presented to the visual attention model. From Table~\ref{table:mnist}, we see that the multi-level model with fast weights reaches a very similar performance to the ConvNet model without requiring any biologically implausible weight sharing.

\subsection{Facial expression recognition} \label{section:faces}

To further investigate the benefits of using fast weights in the multi-level visual attention model, we performed facial expression recognition tasks on the CMU Multi-PIE face database \citep{gross2010multi}. The dataset was preprocessed to align each face by eyes and nose fiducial points. It was downsampled to $48 \times 48$ greyscale. The full dataset contains 15 photos taken from cameras with different viewpoints for each illumination $\times$ expression $\times$ identity $\times$ session condition. We used only the images taken from the three central cameras corresponding to $-15^\circ, 0^\circ, 15^\circ$ views since facial expressions were not discernible from the more extreme viewpoints. The resulting dataset contained $> 100,000$ images. 317 identities appeared in the training set with the remaining 20 identities in the test set. 

\begin{figure}[t]
    \centering
    \vspace{-0.1in}
    \includegraphics[width=10cm]{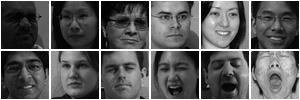}
    ~\\[0.1in]
    \caption{\label{fig:face} \small Examples of the near frontal faces from the MultiPIE dataset.}
\end{figure}
\begin{table}\vspace{-0.in}
\begin{tabular}{lllll}
\toprule
&IRNN&LSTM&ConvNet&Fast Weights\\
\toprule
Test accuracy&81.11&81.32&88.23&86.34\\
\bottomrule
\end{tabular}
\caption{ \label{table:face} \small Classification accuracy comparison on the facial expression recognition task. }
\end{table}
Given the input face image, the goal is to classify the subject's
facial expression into one of the six different categories: neutral,
smile, surprise, squint, disgust and scream. The task is more
realistic and challenging than the previous MNIST
experiments. Not only does the dataset have unbalanced numbers of
labels, some of the expressions, for example squint and disgust, are
are very hard to distinguish. In order to perform well on this
task, the models need to generalize over different lighting
conditions and  viewpoints. We used the same multi-level
attention model as in the MNIST experiments with 200 recurrent hidden
units. The model sequentially attends to non-overlapping 12x12 pixel
patches at two different scales and there are, in total, 24 glimpses.
Similarly, we designed a two layer ConvNet that has a 12x12 receptive
fields.   

From Table~\ref{table:face}, we see that the multi-level fast weights model that knows when to store information outperforms the LSTM and the IRNN.  The results are consistent with previous MNIST experiments. However, ConvNet is able to perform better than the multi-level attention model on this near frontal face dataset. We think the efficient weight-sharing and architectural engineering in the ConvNet combined with the simultaneous availability of all the information at each level of processing allows the ConvNet to generalize better in this task. Our use of a rigid and predetermined policy for where to glimpse eliminates one of the main potential advantages of the multi-level attention model: It can process informative details at high resolution whilst ignoring most of the irrelevant details.  To realize this advantage we will need to combine the use of fast weights with the learning of complicated policies.

\subsection{Agents with memory} \label{section:agents}

While different kinds of memory and attention have been studied extensively in the supervised learning setting~\citep{lstm-handwriting,recurrent-attention,neural-machine-translation}, the use of such models for learning long range dependencies in reinforcement learning has received less attention.

We compare different memory architectures on a partially observable variant of the game "Catch" described in~\citep{recurrent-attention}.
The game is played on an $N\times N$ screen of binary pixels and each episode consists of $N$ frames.
Each trial begins with a single pixel, representing a ball, appearing somewhere in the first row of the column and a two pixel "paddle" controlled by the agent in the bottom row.
After observing a frame, the agent gets to either keep the paddle stationary or move it right or left by one pixel.
The ball descends by a single pixel after each frame.
The episode ends when the ball pixel reaches the bottom row and the agent receives a reward of $+1$ if the paddle touches the ball and a reward of $-1$ if it doesn't.
Solving the fully observable task is straightforward and requires the agent to move the paddle to the column with the ball.
We make the task partially-observable by providing the agent blank observations after the $M$th frame.
Solving the partially-observable version of the game requires remembering the position of the paddle and ball after $M$ frames and moving the paddle to the correct position using the stored information.

\begin{figure}[t]
\vspace{-0.2in}
\subfigure[]{\includegraphics[width = 0.15\linewidth, trim = 0cm -0.2cm 0cm 0cm]{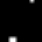}}
\subfigure[]{\includegraphics[width = 0.38\linewidth]{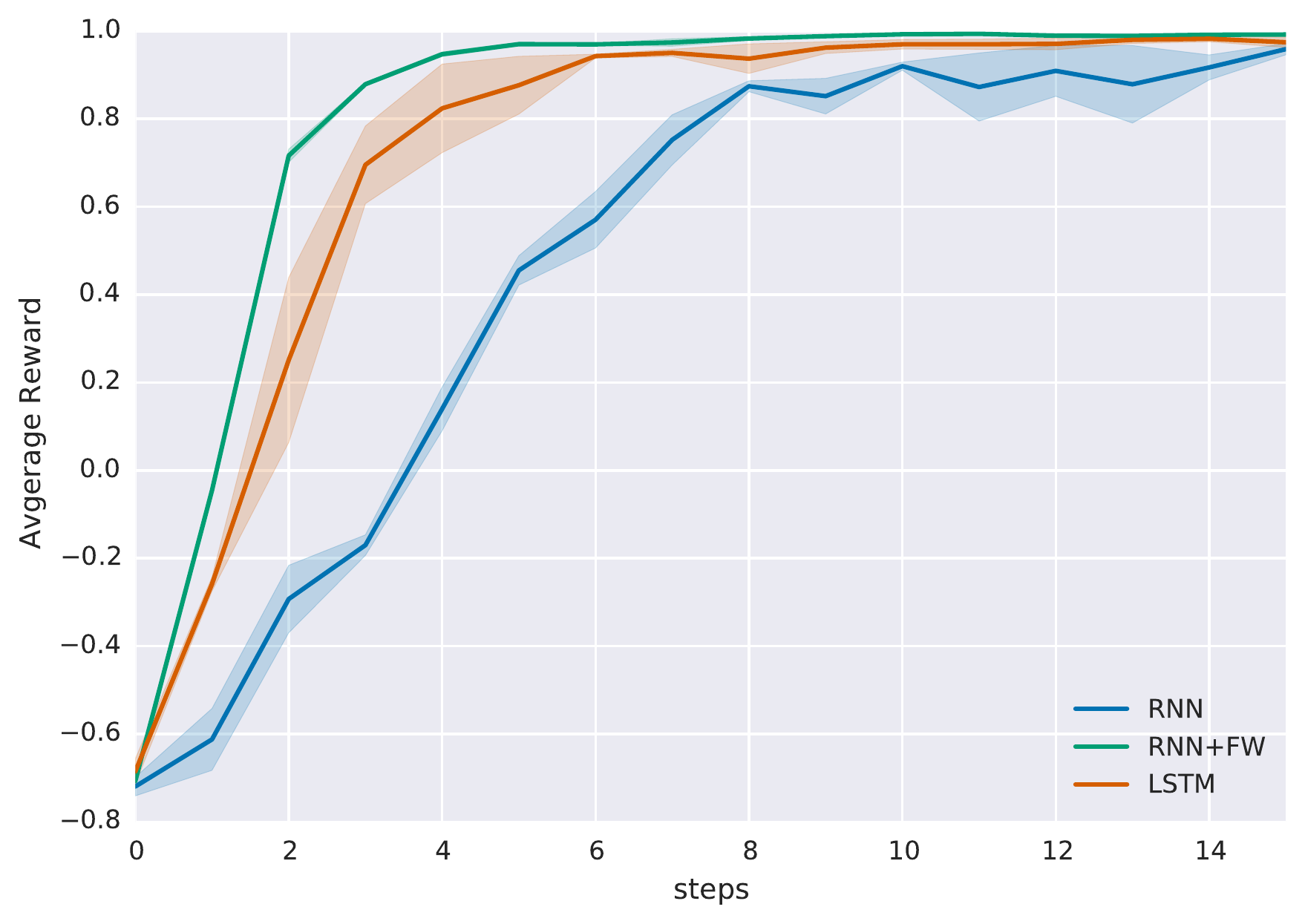}} 
\subfigure[]{\includegraphics[width = 0.38\linewidth]{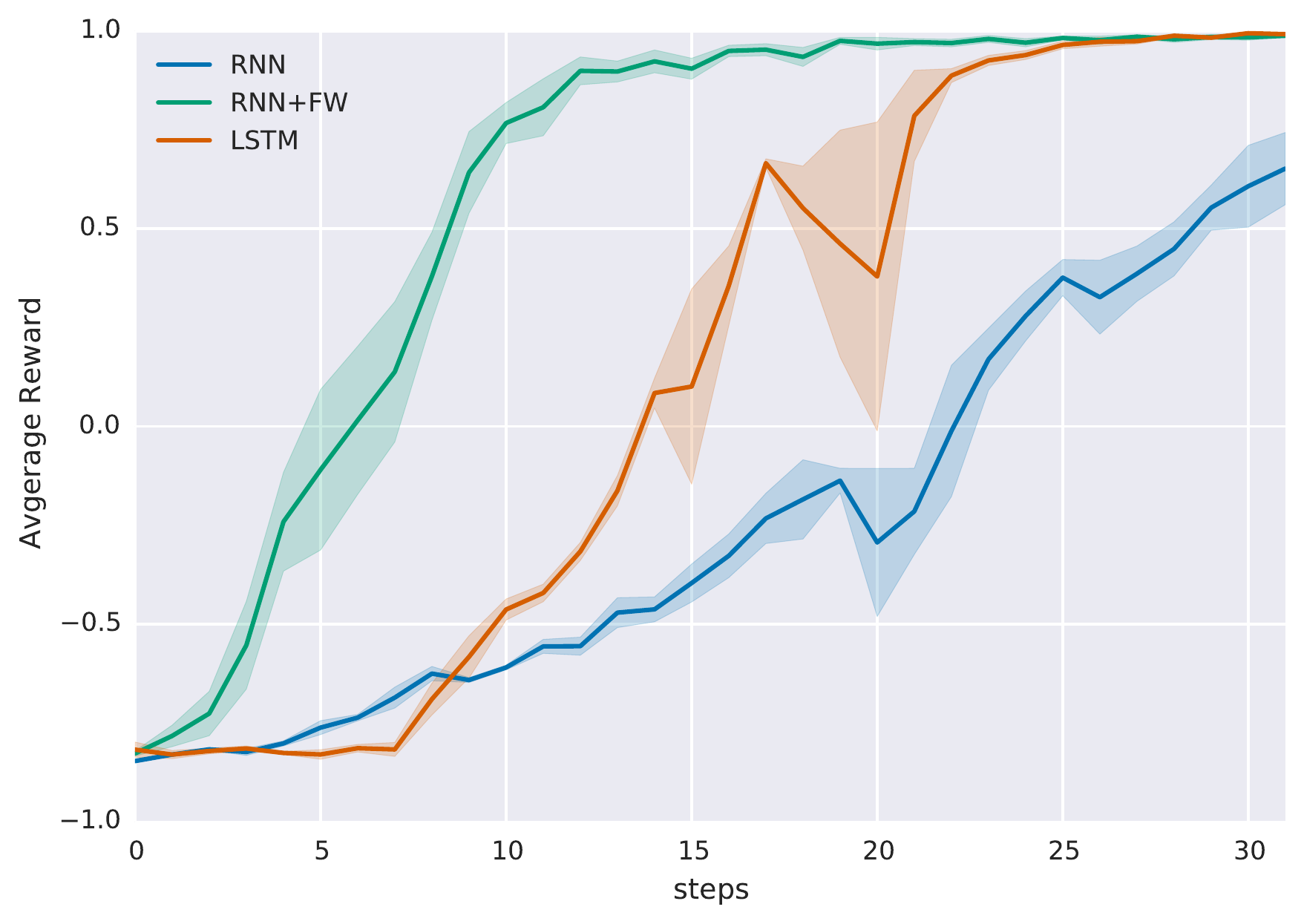}} 
    \vspace{-0.1in}
\caption{a) Sample screen from the game "Catch" b) Performance curves for Catch with $N=16,M=3$.  c) Performance curves for Catch with $N=24,M=5$.}
\label{fig:rlresults}
\end{figure}

We used the recently proposed asynchronous advantage actor-critic method~\citep{asyncrl} to train agents with three types of memory on different sizes of the partially observable Catch task. The three agents included a ReLU RNN, an LSTM, and a fast weights RNN.
Figure~\ref{fig:rlresults} shows learning progress of the different agents on two variants of the game $N=16,M=3$ and $N=24,M=5$.
The agent using the fast weights architecture as its policy representation (shown in green) is able to learn faster than the agents using ReLU RNN or LSTM to represent the policy.
The improvement obtained by fast weights is also more significant on the larger version of the game which requires more memory.

\section{Conclusion}

This paper contributes to machine learning by showing that the performance of RNNs on a variety of different tasks can be improved by introducing a mechanism that allows each new state of the hidden units to be attracted towards recent hidden states in proportion to their scalar products with the current state. Layer normalization makes this kind of attention work much better.  This is a form of attention to the recent past that is somewhat similar to the attention mechanism  that has recently been used to dramatically improve the sequence-to-sequence RNNs used in machine translation.  
The paper has interesting implications for computational neuroscience and cognitive science. The ability of people to recursively apply the very same knowledge and processing apparatus to a whole sentence and to an embedded clause within that sentence or to a complex object and to a major part of that object has long been used to argue that neural networks are not a good model of higher-level cognitive abilities. By using fast weights to implement an associative memory for the recent past, we have shown how the states of neurons could be freed up so that the knowledge in the connections of a neural network can be applied recursively. This overcomes the objection that these models can only do recursion by storing copies of neural activity vectors, which is biologically implausible.

\begin{footnotesize}
  \setlength{\bibsep}{5pt plus 2.0ex}
\bibliography{fw_biblio}
\end{footnotesize}

\appendix
\addcontentsline{toc}{section}{Supplementary Material}
\section*{Supplementary Material}

\section{Experimental details}
\subsection{Associative retrieval}

We used a single hidden layer recurrent neural network which takes a 100 dimensional embedding vector as its input. We compared the fast weights memory against three other different RNN architecture: IRNN, standard LSTM and associative LSTM. The non-recurrent slow recurrent weights are initialized from uniform distribution between $(-1/\sqrt{H}, 1/\sqrt{H})$, where $H$ is the number outgoing weights. The slow weights learning rate is tuned using the 10,000 validation examples. 

\vspace{0.1in}

Below, we provide the specific hyper-parameter settings for the models used in the experiments:

\textbf{Fast weights}: The fast weights learning rate, $\eta$, is set to 0.5 and the fast weights decay rate, $\lambda$, is set to 0.9. The fast weights are updated once at every time step. We experimented with more iterations for the ``inner loop'' and the performance are similar. The recurrent slow weights are initialized to an identity matrix scaled by $0.05$. We use the ReLU activation for $f(\cdot)$ in the recurrent layer.

\vspace{0.1in}

\textbf{IRNN}: The recurrent slow weights are initialized to an identity matrix scaled by $0.5$. ReLU is used as the non-linearity in the recurrent layer.

\vspace{0.1in}

\textbf{Associative LSTM}: We used 4 copies of memory cells for the associative LSTM. There are 3 read-write heads used for storage and retrieval memory access.

\subsection{Integrating glimpses in visual attention models: MNIST and Facial expression recognition}

Both tasks used the similar parameter initialization and the hyper-parameter setup that are comparable to the associative retrieval task mentioned above.

\subsection{Agents with memory}

All agents used recurrent networks to represent the policy. At each time step
the input was passed through a hidden layer with 128 ReLU units and then passed
to the recurrent core. All agents used 128 recurrent cells. The output at every
step was a softmax over the valid actions and a single linear output for the
estimate of the value function. We used random search to find hyperparameters
values for the learning rate, the number of Hebbian steps, and fast weight
learning rate and decay where applicable.  We averaged results over the top 5
models.

\section{Implementing the fast weights ``inner loop'' in biological neural networks}

We considered two different ways of performing this inner loop settling. In method 1 (which is what we use) the inputs to the hidden units after an outer loop transition using W are stored and provide sustained “boundary conditions” during the inner loop settling. In method 2 (which is more biologically plausible)  we simply add the identity matrix to the fast weight matrix so that the inner loop settling tends to sustain the hidden activity vector. For ReLUs, these two methods are equivalent when the fast weight matrix is zero . They are similar but not exactly equivalent when the fast weights are non-zero. Using layer normalization, we found that method 1 worked slightly better than Method 2, but Method 2 would be much easier to implement in a biological network.

\end{document}